\documentclass{article}
\usepackage{ijcai18}

\usepackage{times}
\usepackage{xcolor}
\usepackage{soul}
\usepackage[utf8]{inputenc}
\usepackage[small]{caption}
\usepackage{subfigure}
\usepackage{float}
\usepackage{xcolor}
\usepackage{amsmath}

\usepackage{relsize}
\usepackage{array}
\usepackage{algorithm}
\usepackage{algorithmic}
\usepackage{xcolor}
\usepackage{url}
\usepackage{multirow}
\usepackage{comment}
\usepackage[draft]{todonotes}   % notes showed
\usepackage{subfigure}

\usepackage{amsfonts}
\makeatletter
\def\amsbb{\use@mathgroup \M@U \symAMSb}
\makeatother
\DeclareSymbolFont{bbold}{U}{bbold}{m}{n}
\DeclareSymbolFontAlphabet{\mathbbold}{bbold}
\DeclareMathOperator*{\argmax}{arg\,max}

\specialcomment{sebcomment}{\begingroup\color{blue}\textbf{Seb:}}{\endgroup}
\excludecomment{sebcomment}

\title{COBRAS: Fast, Iterative, Active Clustering with Pairwise Constraints}

\author{Toon Van Craenendonck, Sebastijan Duman\v{c}i\'{c}, Elia Van Wolputte \and Hendrik Blockeel\\ 
Department of Computer Science, KU Leuven, Belgium  \\
\{firstname.lastname\}@kuleuven.be}

\begin{document}

\maketitle

\begin{abstract}
 Constraint-based clustering algorithms exploit background knowledge to construct clusterings that are aligned with the interests of a particular user. This background knowledge is often obtained by allowing the clustering system to pose pairwise queries to the user: should these two elements be in the same cluster or not? Active clustering methods aim to minimize the number of queries needed to obtain a good clustering by querying the most informative pairs first. Ideally, a user should be able to answer a couple of these queries, inspect the resulting clustering, and repeat these two steps until a satisfactory result is obtained. We present COBRAS, an approach to active clustering with pairwise constraints that is suited for such an interactive clustering process. A core concept in COBRAS is that of a super-instance: a local region in the data in which all instances are assumed to belong to the same cluster. COBRAS constructs such super-instances in a top-down manner to produce high-quality results early on in the clustering process, and keeps refining these super-instances as more pairwise queries are given to get more detailed clusterings later on. We experimentally demonstrate that COBRAS produces good clusterings at fast run times, making it an excellent candidate for the iterative clustering scenario outlined above.
\end{abstract}

\section{Introduction}

Clustering is inherently subjective \cite{Caruana06metaclustering,ScienceOrArt}: different users often require very different clusterings of the same dataset, depending on their prior knowledge and goals. 
Constraint-based (or semi-supervised) clustering methods are able to deal with this subjectivity by taking a limited amount of user feedback into account. 
Often, this feedback is given in the form of pairwise constraints  \cite{Wagstaff01constrainedk-means}. The algorithm has no direct access to the cluster labels in a target clustering, but it can perform pairwise queries to answer the question: \textit{do instances $i$ and $j$ have the same cluster label in the target clustering?}
A must-link constraint is obtained if the answer is yes, a cannot-link constraint otherwise.

An effective constraint-based clustering system should satisfy three requirements. 
First, it should allow for an iterative clustering process. In each iteration the user answers several pairwise queries, resulting in pairwise constraints. The clustering system uses these to improve the current clustering. 
This process is repeated until the user is satisfied with the given clustering. 
Second, it should produce high-quality solutions given only a limited number of pairwise queries. 
This motivates the use of active query selection in clustering, in which the clustering system tries to determine the most informative queries. 
Third, the process should be fast. 
The workflow described above is inherently interactive: the user repeatedly answers pairwise queries and inspects the updated clustering. 
For this to work in practice, both producing the clusterings and deciding on which pairs to query next should be fast.

None of the existing constraint-based clustering systems fulfills all of these requirements.
First, most of them assume that all pairwise constraints are given prior to running the clustering algorithm \cite{xing2002distance,probframework,Bilenko2004,Mallapragada2008}, which makes them non-iterative. 
Second, traditional systems typically query random pairs \cite{xing2002distance,Bilenko2004}, which might not be the most informative ones and result in low-quality solutions. Several active constraint-based clustering methods have been proposed \cite{basu:sdm04,Mallapragada2008} that outperform random query selection, but most of them are non-iterative. NPU \cite{Xiong2014} is an example of a clustering framework that does satisfy the first two requirements. However, it does not satisfy the third as it requires re-clustering the entire dataset after every few constraints, which becomes prohibitively slow for large datasets. 

The approach closest to fulfilling the requirements outlined above is COBRA \cite{COBRA}, a recently introduced method based on the concept of super-instances.
A super-instance is a set of instances that are assumed to belong to the same cluster in the unknown target clustering.
COBRA consists of two steps: it first over-clusters the data using K-means to construct these super-instances and then merges them into clusters based on pairwise constraints.
COBRA was shown to produce high-quality clusterings at fast run times. However, a fixed number of super-instances has to be specified prior to the clustering process.
Using a small number of super-instances results in good high-level clusterings using few queries, but these clusterings cannot be refined as more queries are answered. 
If the number of super-instances is large, more fine-grained structure can be found as more queries are answered, but at the cost of lower quality clusterings early on in the clustering process. Hence, more queries are needed before a good result is obtained.

In this work, we introduce COBRAS (for Constraint-based Repeated Aggregation and Splitting), an active clustering system satisfying all the requirements outlined above.
In contrast to COBRA, it does not need a fixed set of super-instances.
Instead, it combines the \textit{bottom-up procedure} of merging super-instances with an incremental \textit{top-down search} for good super-instances. 
By doing this it largely mitigates the trade-off present in COBRA: in the beginning the number of super-instances is small which allows getting a reasonable coarse-grained clustering using few queries; as more queries are answered these super-instances are refined, allowing to capture more fine-grained structure.

The remainder of this paper is structured as follows. In the next section, we describe existing work on constraint-based clustering. Next, we discuss COBRAS in more detail and give an algorithmic description. In our experimental section we demonstrate that COBRAS is the most suitable clustering method to be used in the iterative workflow described above, as it produces high-quality clusterings at fast run times. 

\section{Related work}

The most common way to develop a constraint-based clustering method is to extend an existing unsupervised method. 
One can either adapt the clustering procedure to take the pairwise constraints into account \cite{Wagstaff01constrainedk-means,Rangapuram2012,Wang2014}, or use the existing procedure with a new similarity metric that is learned based on the constraints \cite{xing2002distance,Davis:2007:IML:1273496.1273523}. 
Alternatively, one can also modify both the similarity metric and the clustering procedure \cite{Bilenko2004,probframework}.

Traditional constraint-based clustering methods assume that a set of constraints is given, and in practice this set is often obtained by querying random pairs. Basu et al.\ \shortcite{basu:sdm04} introduce active constraint selection and show that selecting a set of informative queries can outperform querying random pairs. 
In their method, the entire set of constraints is queried prior to a single run of the constraint-based clustering algorithm. 
Xiong et al.\ \shortcite{Xiong2014} introduce NPU, an active selection procedure in which the data is clustered multiple times and each resulting clustering is used to determine which pairs to query next based on the principle of uncertainty sampling.   

COBS \cite{COBS} is quite different from the previously discussed methods: it uses pairwise constraints to select and tune an unsupervised clustering algorithm. COBS generates a large set of clusterings by varying the hyperparameters of several unsupervised clustering algorithms, and selects the clustering from the resulting set that satisfies the most pairwise constraints. 

COBRA \cite{COBRA} is a recently proposed method that is inherently active: deciding which pairs to query is part of its clustering procedure. First, COBRA uses K-means to cluster the data into super-instances. The number of super-instances, denoted as $N_S$, is an input parameter. Initially, each of the super-instances forms its own cluster. In the second step COBRA repeatedly queries the pairwise relation between the closest pair of (partial) clusters between which the relation is not known yet and merges clusters if necessary, until all relations between clusters are known. 

\begin{figure}[ht]
\centering
  \centering
  \includegraphics[width=1.0\linewidth]{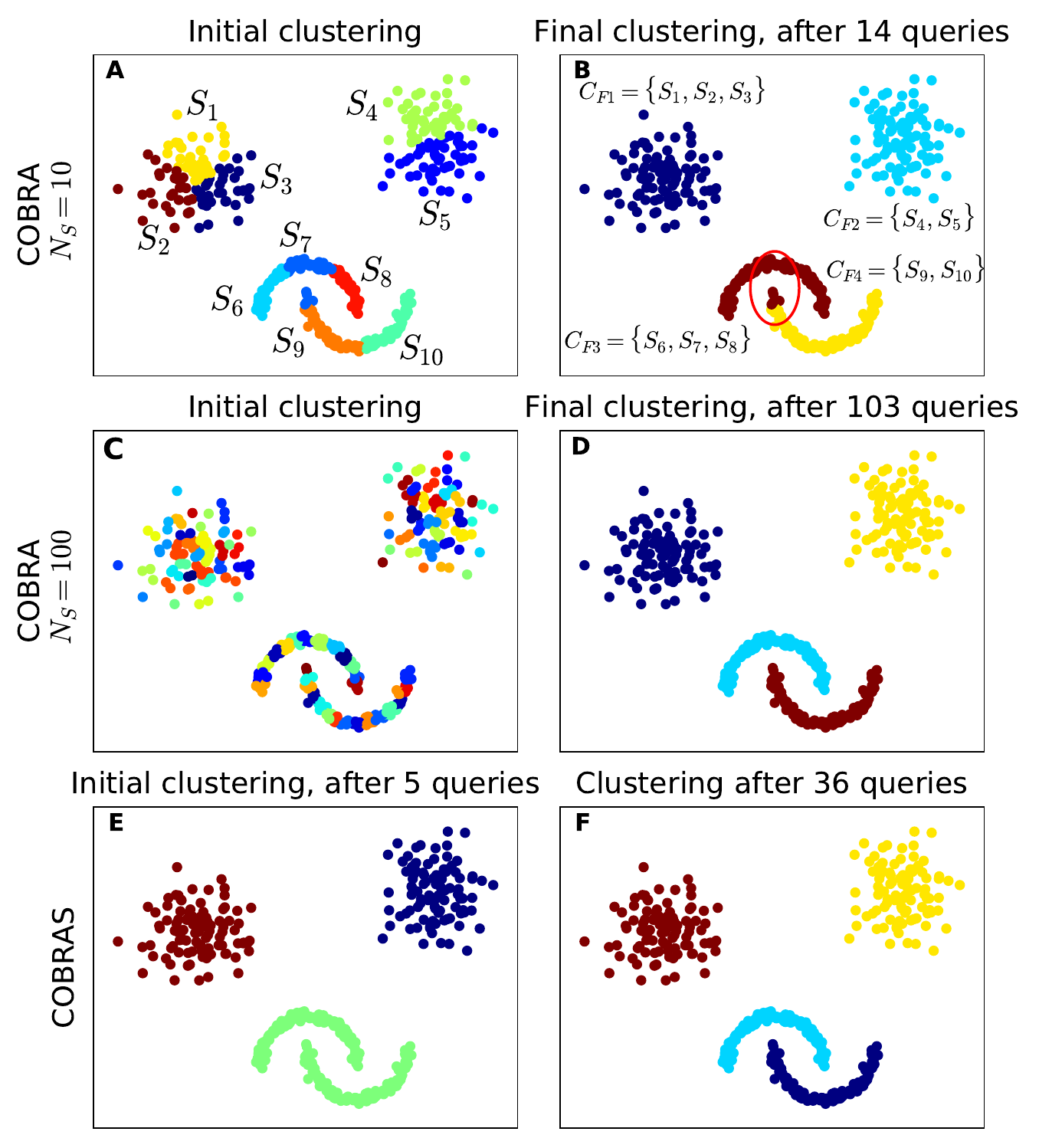}
  \caption{A: The starting situation of COBRA with 10 super-instances (COBRA-10). Initially, each cluster consists of a single super-instance. B: final result of COBRA-10. Each of the clusters is represented as a set of super-instances. The final clustering is not correct, as $S_7$ contains instances from two actual clusters. C: The initial solution of COBRA-100, which is highly over-clustered. D: the final clustering of COBRA-100. E: the clustering produced by COBRAS after 5 queries. The first two queries are used to determine the initial splitting level (which was $k=2$), the next three for determining the pairwise relation between the first three super-instances. F: after 36 queries, COBRAS produces the correct clustering.  }
  \label{fig:example}
\end{figure}%

The results of COBRA were found to be strongly dependent on the number of super-instances $N_S$. A small value of $N_S$ has the advantage that is gives clusterings of reasonably good quality given few pairwise queries, but lacks the possibility of getting more fine-grained results. A large value of $N_S$ typically results in higher quality clusterings, but these clusterings only appear after answering a relatively high number of queries. This is illustrated for a toy dataset in Figure \ref{fig:example}a-d: 10 super-instances is not enough to get a correct clustering (the incorrectly clustered part is marked with a red ellipse), 100 super-instances result in a correct clustering, but only after 103 queries are answered. Note that this problem cannot be solved by tuning $N_S$: there is no value of $N_S$ that produces both a decent high-level clustering given few constraints, and a fine-grained clustering given more constraints.  

\section{COBRAS: Constraint-based Repeated Aggregation and Splitting}

The key problem when running COBRA with a small $N_S$ is that super-instances often contain instances from different clusters (this happens e.g.\ for $S_7$ in Figure \ref{fig:example}a). COBRA cannot assign all of these instances to the correct clusters, as each super-instance is treated as a single unit.

COBRAS solves this problem by allowing super-instances to be refined. It starts with a single super-instance that contains all instances, and repeatedly refines this super-instance until a satisfactory clustering is obtained. More specifically, each iteration of COBRAS consists of two steps. First, it removes the largest super-instance from its cluster and splits it into two new super-instances. Second, it determines the relation of these two new super-instances to the existing clusters by running the merging step of COBRA on the new set of super-instances. By using this procedure of refining super-instances, COBRAS uses a small number of super-instances in the beginning of the clustering process, and a larger number as more queries are answered. This allows it to both produce reasonable high-level clusterings early on, and more fine-grained ones later. Panels (e) and (f) in Figure \ref{fig:example} illustrate the initial and final clusterings produced by COBRAS, and shows that it indeed performs well for both a small and larger number of queries.

\begin{sebcomment}
\textit{I'm wondering now - why not put the previous two paragraphs in caption? The caption as it is now is quite useless - one look at the image and you know what's on. However, it doesn't tell the main message; that's in the text. But again, when you read the text the picture is either on the other page so you have to keep scrolling back and forth; if you print it out, you still have to constantly go between text on one page and picture on different page. If this description is in the caption, then everything that is relevant for the picture is on the same place. I think only the paragraph below is enough for the text itself, everything else seem to better fit in the caption.}
\end{sebcomment}

\begin{comment}
\textcolor{blue}{\textbf{Seb:} too much space spent on COBRA, approx the same as for COBRAS?}
\textcolor{red}{\textbf{Toon:} still spending much time on the example, but now more explicitly framed as motivation for COBRAS}
\end{comment}

\begin{sebcomment}
\textit{Is COBRAS part of the picture actually necessary? The picture is supposed to show weaknesses of COBRA, so from that perspective I think that COBRAS part can be left out - it doesn't bring anything to support that, but takes quite some space. It can be mentioned in the text that COBRAS achieve equal performance but with smaller number of constraints. Maybe an intermediate COBRA-40/50/60 step might be more interesting showing how it improves a bit by bit.}
\end{sebcomment}

\subsection{Algorithmic description}
COBRAS is described in Algorithm \ref{algo:cobras}. In this algorithm a super-instance $S$ is a set of instances, a cluster $C$ is a set of super-instances, and a clustering $\mathcal{C}$ is a set of clusters. COBRAS starts with a single super-instance $S$ that contains all instances, which constitutes the only cluster $C$ (line 2). As long as the user keeps answering queries, COBRAS keeps refining the set of super-instances and the corresponding clustering (lines 3-10). In each iteration it selects the largest super-instance (line 4) and determines an appropriate splitting level for this super-instance (line 5, this is detailed in Algorithm \ref{algo:determine_k} which is discussed in the next subsection). COBRAS splits the selected super-instance $S_{split}$ into $k$ new super-instances by clustering the instances in $S_{split}$ using K-means (line 6). $S_{split}$ is then removed from its original cluster (line 7), and a new cluster is added for each of the newly created super-instances (line 8). Finally, in the last step of the while iteration COBRA is used to determine the pairwise relations between the newly added clusters (which each consist of a single super-instance), and the existing ones. 

\begin{comment}
If this should be included, do it when discussing COBRA in the example.
The COBRA procedure used in line 8 of the algorithm differs from COBRA as it is defined in \cite{COBRA} in the sense that it already starts from a partial solution. With this we mean that several clusterings in $\mathcal{C}$ are already the result of merging super-instances in previous iterations. Further, the pairwise relation between many clusters in $\mathcal{C}$ might already be known prior to the COBRA call in line 8: of course the new COBRA run will not re-query these relations (for this purpose $ML$ and $CL$ are passed in the call), it will only determine the relations of the newly created clusters to the existing ones. 
\end{comment}

\begin{algorithm}[ht]
\caption{COBRAS}
\label{algo:cobras}
\begin{algorithmic}[1]
 \REQUIRE $\mathcal{X}$: a dataset\\
 	\ \ \ \ \ \ \ \ \ \ \   $q$: a query limit
\ENSURE $\mathcal{C}$: a clustering of $D$
\STATE $ML = \emptyset, CL = \emptyset$
\STATE $S = \{\mathcal{X}\}, C = \{S\}, \mathcal{C} = \{C\}$
\WHILE {$|ML| + |CL| < q$}
	\STATE {$S_{split}, C_{origin} =  \argmax_{S \in C, C \in \mathcal{C}}{|S|}$}
    \STATE {\small {$k, ML, CL = \texttt{determineSplitLevel}(S_{split}, ML, CL)$} }
    \STATE {$S_{new_1}, \ldots, S_{new_k} = \texttt{K-means}(S_{split},k)$}
    \STATE {$C_{origin} = C_{origin} \setminus \{S_{split}\}$}
    \STATE {$\mathcal{C} = \mathcal{C} \cup \{ \{ S_{new_1} \}, \ldots, \{ S_{new_k} \}  \}$}
    \STATE {$\mathcal{C}, ML, CL = \texttt{COBRA}(\mathcal{C},ML,CL)$}
\ENDWHILE
\RETURN $\mathcal{C}$
\end{algorithmic}    
\end{algorithm}

\begin{algorithm}[ht]
\caption{\emph{determineSplitLevel}}
\label{algo:determine_k}
\begin{algorithmic}[1]
 \REQUIRE $\mathcal{S}$: a set of instances that is to be split
\ENSURE $k$: an appropriate splitting level\\
\ \ \ \ \ \ \ \    $ML$: the obtained ML constraint\\
\ \ \ \ \ \ \ \   $CL$: the obtained CL constraints
\STATE $d = 0$, $ML = \emptyset$, $CL = \emptyset$
\WHILE {no must-link obtained}
	\STATE{$\mathcal{S}_1$, $\mathcal{S}_2$ = k-means($\mathcal{S}$,2)}
    \IF{must-link(medoid($\mathcal{S}_1$), medoid($\mathcal{S}_2$))}
    	\STATE {add (medoid($\mathcal{S}_1$), medoid($\mathcal{S}_2$)) to $ML$}
        \STATE {$d = \max(d,1)$}
    	\RETURN $2^d$, $ML$, $CL$
     \ELSE
      	\STATE {add (medoid($\mathcal{S}_1$), medoid($\mathcal{S}_2$)) to $CL$}
     	\STATE $\mathcal{S} =$ pick between $\mathcal{S}_1$ and $\mathcal{S}_2$ randomly
        \STATE {$d\texttt{++}$}
      \ENDIF
     
\ENDWHILE
\end{algorithmic}    
\end{algorithm}

\begin{figure*}[ht]
\centering
  \centering
  \includegraphics[width=0.752\linewidth]{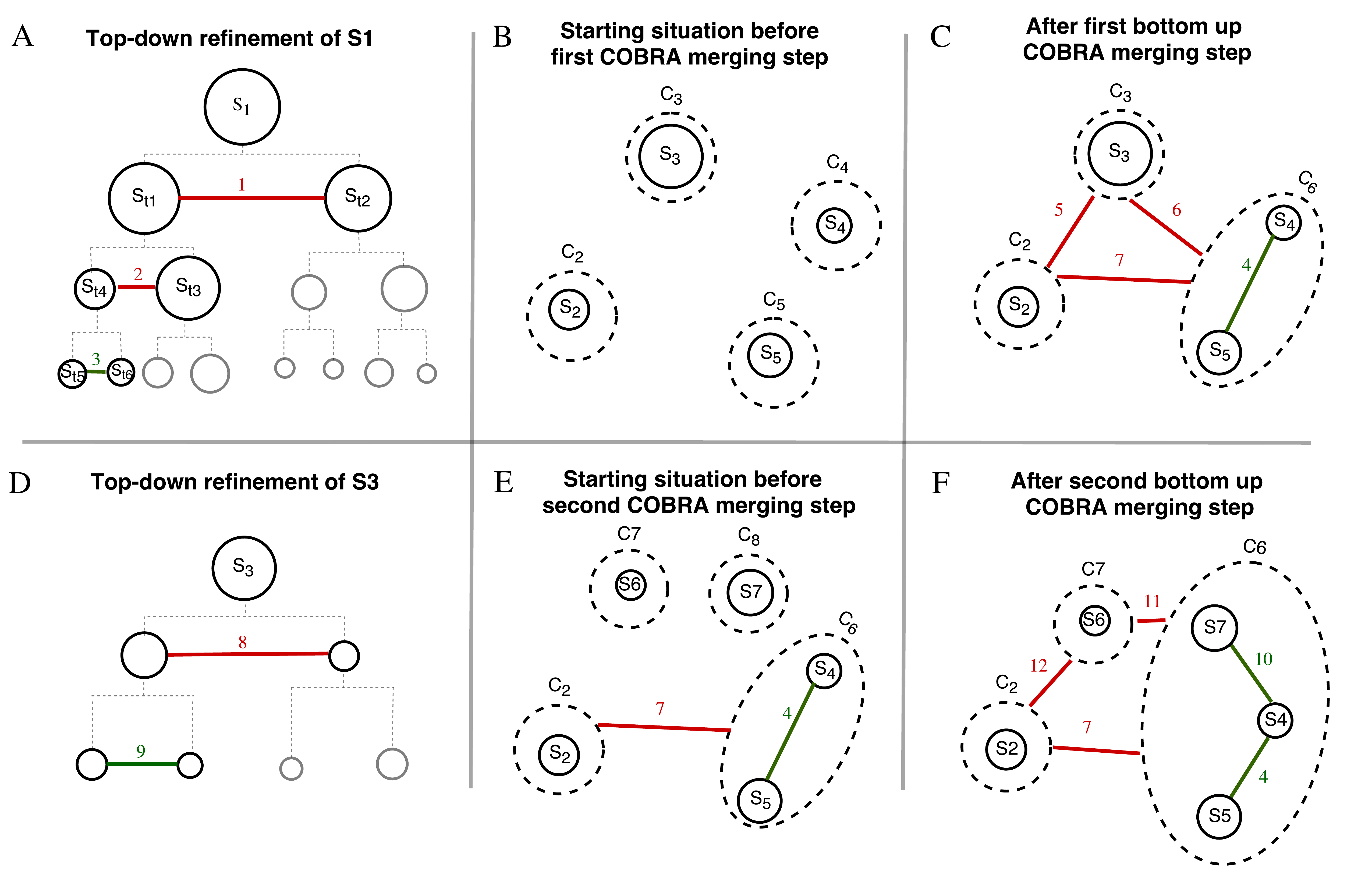}
  \caption{(a) COBRAS decides to split the initial super-instance $S_1$ into 4 new ones, as discussed in section \ref{sec:splittinglevel}. (b) $S_1$ was removed from the set of clusters (rendering it empty), and a new cluster was added for each of the newly created super-instances. This is the starting situation for the first bottom-up COBRA run. (c) The results of the COBRA run. COBRA queries the relations between the closest pair of clusters between which the relation is not known yet, until all relations are known. For example, it started by querying the relation between $S_4$ and $S_5$ by querying the pairwise relationship between their medoids. This resulted in a must-link constraint, and in the merging of $C_4$ and $C_5$ into $C_6$. (d) At the beginning of a new iteration, COBRAS selects the largest super-instance ($S_3$ in this case) to refine it further. In this case, $S_3$ is split into 2 new super-instances. (E) This results in two new clusters containing these super-instances before the start of a new COBRA merging step. (F) The pairwise relation between these two new clusters and the existing ones is determined using COBRA, leading to a new clustering. This illustrates the situation at the end of the second COBRAS iteration. }
  \label{fig:algo_steps}
\end{figure*}%

\subsection{Determining the splitting level $k$}
\label{sec:splittinglevel}
\begin{comment}
An important question in each iteration of COBRAS is the number of super-instances $k$ that the current super-instance $S_{split}$ will be split in. An appropriate value of $k$ does not only depend on the data, but also on the user's clustering intent. Consider for example the CMU faces dataset, which consists of images of different people, taking different poses, while expressing different emotions, with or without sunglasses. This dataset can naturally be clustered in multiple ways: one user might want to cluster it by whether the person in the image wears sunglasses or not, another based on the identity of the person in the image. For the former, $k=2$ in the first while iteration would be appropriate, whereas for the latter an initial value around $k=20$ is better suited. From this example, it is clear that the procedure for determining $k$ should take the preferences of the user into account, and cannot depend on data alone. 
\end{comment}
Algorithm \ref{algo:determine_k} describes the procedure that COBRAS uses to determine the splitting level $k$ for a super-instance $S$. The procedure tries to search for a $k$ such that the new super-instances will be pure w.r.t.\ the unknown target clustering. To check the purity of $S$, COBRAS splits it into two new (temporary) super-instances, and queries the relation between their medoids. Obtaining a must-link constraint indicates that the super-instance was pure, and we are at an appropriate level of granularity. Obtaining a cannot-link constraint on the other hand indicates that the original super-instance contained instances that should be in different clusters, and further splitting is warranted. In this case, one of the two new super-instances is split further, until a must-link constraint is obtained. 

We illustrate this procedure based on the example given in Figure \ref{fig:algo_steps}(a), in which the splitting level for the initial super-instance $S_1$ is determined. We first split the super-instance into two new sets of instances, in this case $S_1$ is split into $S_{t1}$ and $S_{t2}$. The $t$ subscript indicates that these sets of instances are only temporary, i.e.\ they are only created in the process of finding the splitting level and afterward discarded. Next, the pairwise relation between the two newly created super-instances is queried. In this case, querying the relation between $S_{t1}$ and $S_{t2}$ results in a cannot-link constraint (indicated by constraint 1 in Figure \ref{fig:algo_steps}(a)). This cannot-link constraint indicates that it is indeed useful to split $S_1$ into smaller super-instances, as it contains elements that should be in different clusters. We repeat this process for $S_{t1}$, which in this case results in a cannot-link constraint between $S_{t3}$ and $S_{t4}$. Again, this indicates the usefulness of further splitting $S_{t1}$ into smaller super-instances. We finally repeat the process for $S_{t3}$, and obtain a must-link constraint between $S_{t5}$ and $S_{t6}$. This indicates that $S_{t3}$ was at an appropriate level of granularity. The algorithm assumes that this level of granularity is also appropriate for the remainder of the instances in $S_1$ (and not only for the single branch that we followed to $S_{t3}$), and determine $k=4$ to be an appropriate splitting level ($S_{t3}$ was at the second level of the tree, hence we split into $2^2$ new super-instances). Line 6 in Algorithm \ref{algo:determine_k} ensures that a super-instance is split into at least two new ones.

The remainder of Figure \ref{fig:algo_steps} illustrates two iterations of the entire COBRAS clustering process.

\begin{sebcomment}
The description is clear as is. It could potentially be improved by stating the general idea first - currently it says what is being done in the manner of the report. 
\end{sebcomment}

\begin{sebcomment}
\textit{same as for COBRA weakness -- put in caption?}
\end{sebcomment}

\section{Experimental evaluation}

\label{sec:results}

In this section, we discuss the experimental evaluation of COBRAS.

\subsection*{Existing Constraint-based Algorithms}
We compare COBRAS to the following state-of-the-art constraint-based clustering algorithms:
\begin{itemize}
	\item \textbf{COBS} \cite{COBS} uses constraints to select and tune an unsupervised clustering algorithm. We use the active variant in our experiments.
    \item \textbf{COBRA} \cite{COBRA} is the algorithm that is most related to COBRAS, as discussed earlier in this paper. We run it with 10, 25 and 50 super-instances.
    
    \item \textbf{NPU} \cite{Xiong2014} is an active constraint selection framework that can be used with any non-active constraint-based clustering method. It constructs neighborhoods of points that are connected by must-link constraints, with cannot-link constraints between the different neighborhoods. It repeatedly selects the most informative instance, and queries its neighborhood membership by means of pairwise constraints. NPU is an iterative method: after neighborhood membership is determined, the data is re-clustered and the obtained clustering is used to determine the next pairwise queries. NPU can be used with any constraint-based clustering algorithm, and we use it with the following two:
    		\begin{itemize}
                \item \textbf{MPCKMeans} \cite{Bilenko2004} is an extension of K-means that exploits constraints through metric learning and a modified objective. We use the implementation in the WekaUT package \footnote{{\scriptsize \url{http://www.cs.utexas.edu/users/ml/risc/code/}}}.
                \item \textbf{COSC} (for Constrained Spectral Clustering) \cite{Rangapuram2012} is an extension of spectral clustering optimizing for a modified objective. We use the code provided by the authors \footnote{{\scriptsize \url{http://www.ml.uni-saarland.de/code/cosc/cosc.htm}}}.
            \end{itemize}
\end{itemize}

COSC-NPU and MPCKMeans-NPU require knowing the number of clusters $K$ prior to clustering, and in our experiments this true $K$ (as indicated by the class labels) is given to these algorithms. 
Note that, in practice, this number K is often not known in advance, and that this constitutes a clear advantage of these algorithms over the others in the experimental evaluation.

\subsection*{Datasets}
We use the same datasets as those used in the evaluation of COBRA \cite{COBRA}. These include the following 14 UCI datasets: iris, wine, dermatology, hepatitis, glass, ionosphere, optdigits389, ecoli, breast-cancer-wisconsin, segmentation, column\_2C, parkinsons, spambase, sonar and yeast. These were selected because of their repeated use in earlier work on constraint-based clustering (for example, \cite{Bilenko2004,Xiong2014}). Optdigits389 contains digits 3, 8 and 9 of the UCI handwritten digits data \cite{Bilenko2004,Mallapragada2008}. Duplicate instances are removed from all of these datasets, and the data is normalized between 0 and 1. Further, we use the CMU faces dataset, containing 624 images of 20 persons with different poses and expressions, with and without sunglasses. This dataset has four natural clustering targets: identity, pose, expression and sunglasses. A 2048-value feature vector is extracted for each of the images using the pre-trained Inception-V3 network \cite{inceptionnet}. Further, two clustering tasks are included for the 20 newsgroups text dataset: clustering documents from 3 newsgroups on related topics (the target clusters are comp.graphics, comp.os.ms-windows and comp.windows.x, as in \cite{basu:sdm04,Mallapragada2008}), and clustering documents from 3 newsgroups on very different topics (alt.atheism, rec.sport.baseball and sci.space, as in \cite{basu:sdm04,Mallapragada2008}). To extract features from the text documents we apply tf-idf, followed by latent semantic indexing (as in \cite{Mallapragada2008}) to reduce the dimensionality to 10. In summary, 17 datasets are used in our experiments, for which 20 clustering tasks are defined (14 UCI datasets, 4 target clusterings for the CMU faces data, and 2 subsets of the newsgroups data).

\begin{figure*}
\centering     %%% not \center
\subfigure[]{\label{fig:rank_comparison}\includegraphics[width=0.48\textwidth]{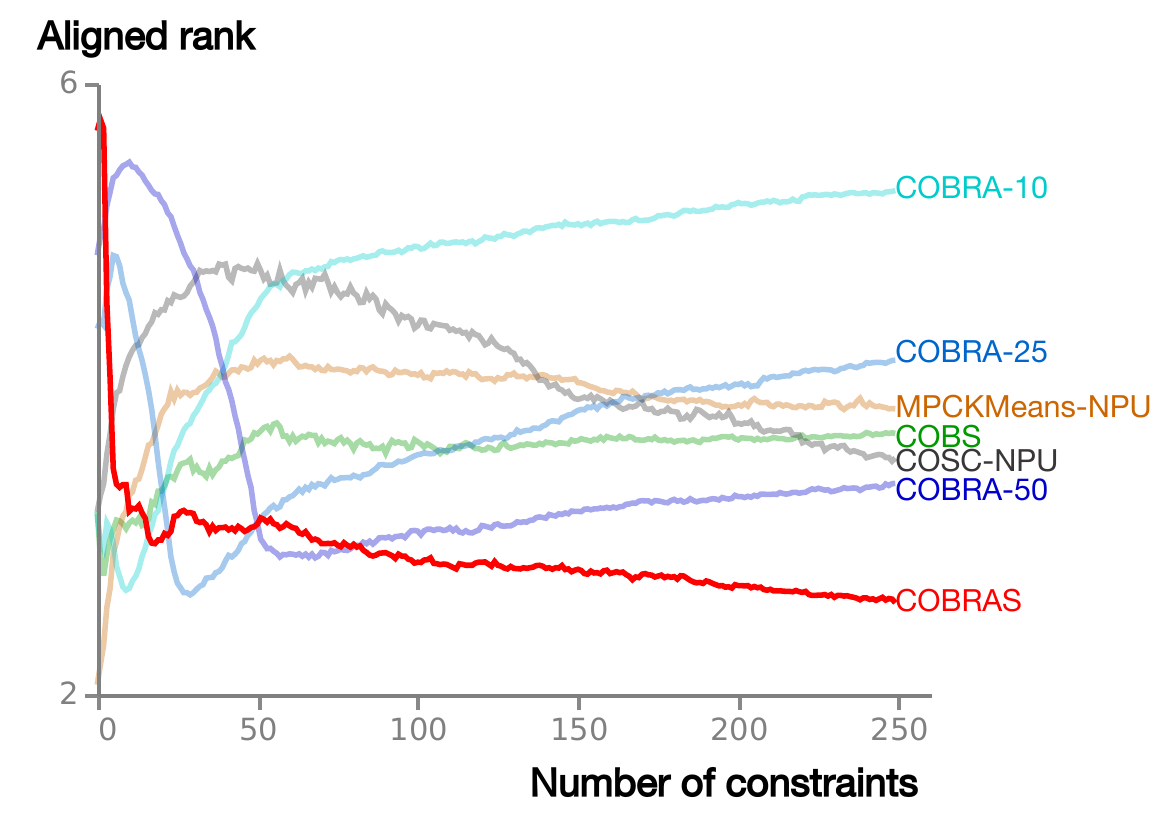}}
\hspace*{25px}
\subfigure[ ]{\label{fig:ari_comparison}\includegraphics[width=0.41\textwidth]{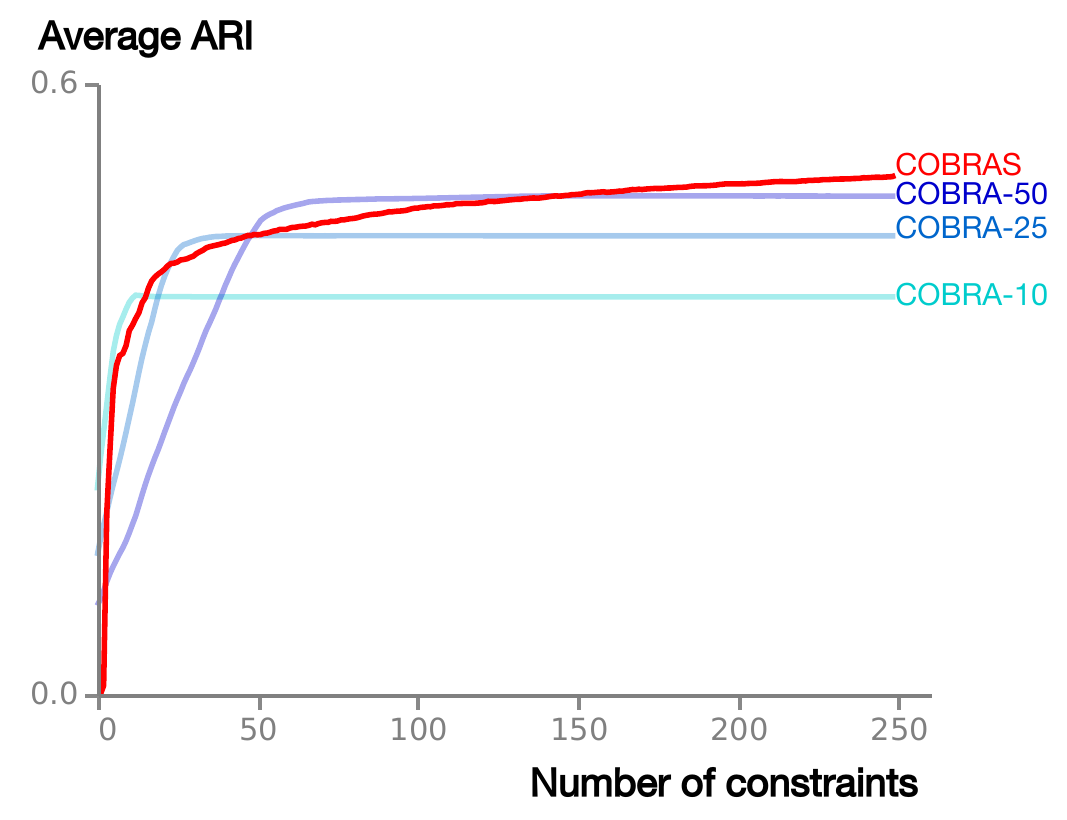}}
\caption{(a) Aligned rank for all methods over all clustering tasks (b) Average ARI of COBRAS and COBRA instantiations over all clustering tasks. The average ARI for other competitors is omitted to not clutter the figure too much.}
\end{figure*}

\subsection*{Experimental methodology}
We perform 10-fold cross-validation 10 times (similar to e.g.\ \cite{basu:sdm04} and \cite{Mallapragada2008}), and report averaged results. The algorithms always cluster the full dataset, but can only query the relations between pairs that are both in the training set. The quality of the resulting clustering is evaluated by computing the Adjusted Rand index (ARI, \cite{ARI}), only on the instances in the test set. The ARI measures the similarity between the produced clusterings and the ground-truth indicated by the class labels. A score of 0 means that the clustering is random, 1 means that it is exactly the same as the ground-truth. The score for an algorithm for a particular dataset is given by the average ARI over the 10 repetitions of 10 fold cross-validation.

We make sure that COBRAS and COBRA do not query any test instances during clustering by only using training instances to compute the medoids of the super-instances. For NPU, pairs involving an instance from the test set are simply excluded from selection. 

In each while iteration of COBRAS, a super-instance is split and COBRA is run on the resulting new set of clusterings. If the user stops answering pairwise queries before the end of the COBRA run (which is simulated frequently in the experiments: we consider the intermediate clusterings after each query), COBRAS returns the clustering as it was at the beginning of the while iteration. The clustering that is returned is only updated after the COBRA run, which prevents us from returning clusterings for which the merging step was not finished yet. This holds for all COBRA runs expect the first one, as in that case there is no real prior clustering at the beginning of the iteration.

COBRA is not able to handle an unlimited amount of pairwise queries: once all the relations between super-instances are known, the clustering process naturally stops. In our experiments, we assume that COBRA simply keeps returning its final clustering after this point, which allows us to compare all algorithms for the same number of pairwise queries.

\subsection*{Clustering quality}

Figure \ref{fig:rank_comparison} shows the aligned ranks for COBRAS and all competitors over all clustering tasks\footnote{For COSC-NPU we set a timeout of 24h for each run of 250 queries for spambase. Typically it only got to 40 queries after that time. We considered the last clustering produced within 24h to be the final one, and use it in the results for all remaining queries in producing the graphs. }. In contrast to the regular rank, the aligned rank \cite{alignedrank,GARCIA20102044} takes the relative differences between algorithms for individual datasets into account. The first step in computing it is to calculate the average ARI achieved for each dataset over all algorithms. Then, for each algorithm the difference between its ARI and this average is calculated, and the resulting differences are sorted from $1$ to $kn$ ($k$ the number of algorithms, $n$ the number of datasets). The aligned rank for an algorithm is then the average of the positions of its entries in the sorted list.  

The figure shows that COBRAS is clearly the best choice for the iterative clustering scenario that was outlined at the beginning of the paper. Only if the user knows in advance how many queries she will answer and does not care about the quality of intermediate results is COBRA the preferred algorithm. In particular, COBRA-10 outperforms COBRAS for 10 queries, COBRA-25 for 25 queries, and COBRA-50 for 50 queries. In many practical applications of clustering, however, the query budget is not known in advance and the quality of intermediate clusterings does matter. None of the COBRA instantiations are suited for this scenario. For example, COBRA-10 performs well for a very small number of queries, but lacks the ability to keep refining clusterings which results in a large performance gap with COBRAS for larger numbers of queries. COBRA-50, on the other hand, is clearly outperformed by COBRAS for a small number of pairwise queries (as it starts from heavily over-clustered solutions). 

A similar argument can be made in the comparison of COBRAS to the other competitors (MPCKMeans-NPU, COSC-NPU and COBS). Furthermore, it is important to realize that COSC-NPU and MPCKMeans-NPU are given the true number of clusters prior to clustering, which explains their good relative performance for a very small number of queries.

\begin{figure}[ht]
\centering
  \centering
  \includegraphics[width=1.0\linewidth]{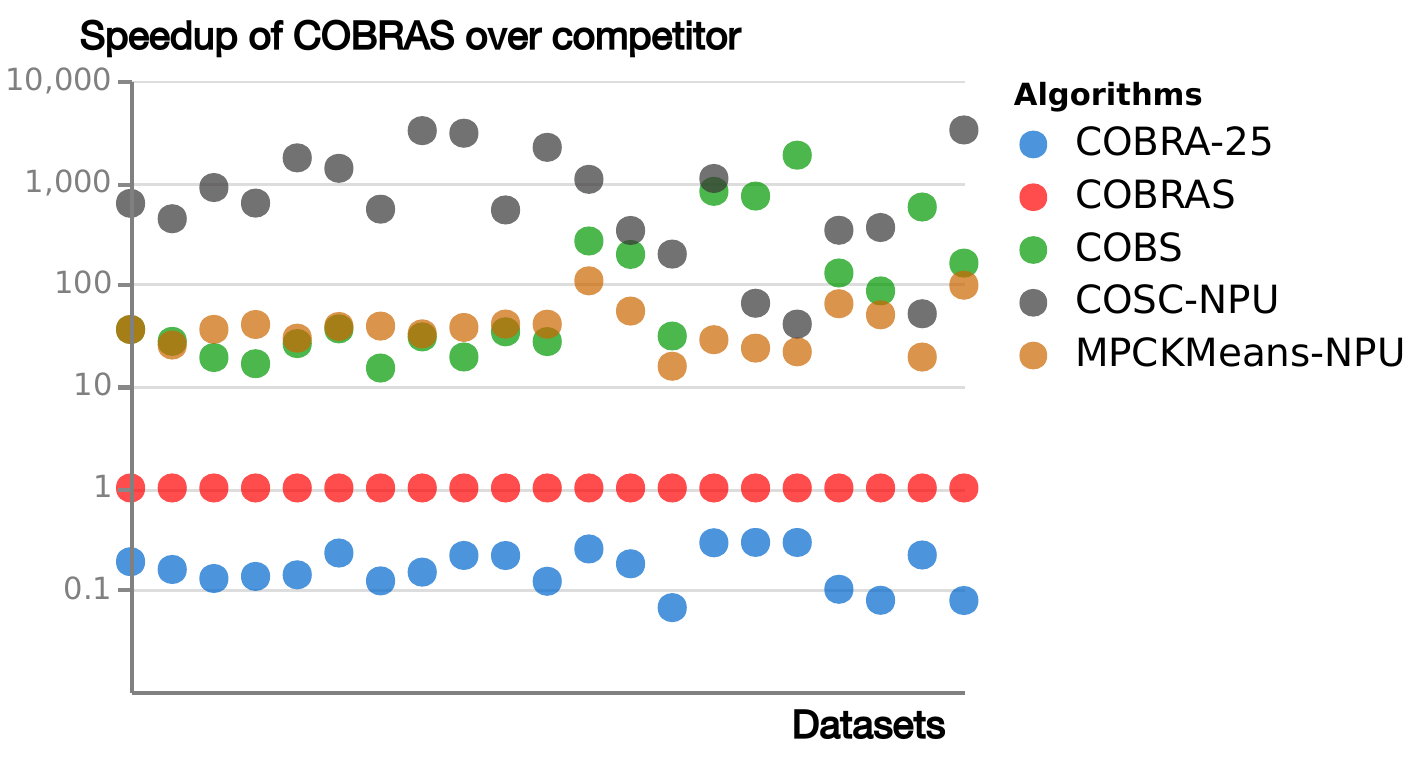}
  \caption{Ratio of COBRAS to competitors run time for 20 clustering tasks. For COBRA we only include the run times of COBRA-25 to not clutter the graph, also the run times for COBRA-10 and COBRA-50 are typically lower than all others.}
  \label{fig:runtimes}
\end{figure}%

Figure \ref{fig:ari_comparison} shows the average ARI of COBRAS and the COBRA instantiations over all clustering tasks. We only show the comparison to the COBRA instantiations to avoid having an overly cluttered figure. The aligned rank comparison has the advantage over the average ARI that it does not depend on immediate comparisons between ARIs on different datasets, but the disadvantage that it does not reflect the actual differences in ARIs between the methods. Figure \ref{fig:ari_comparison} confirms the conclusion drawn from Figure \ref{fig:rank_comparison}: it shows that COBRAS is preferable to each individual COBRA instantiation. It also puts the performance gap between COBRA-25 and COBRAS that Figure \ref{fig:rank_comparison} suggests for 25 constraints into perspective: the aligned rank indicates that COBRA-25 systematically outperforms COBRAS for 25 queries, but Figure \ref{fig:ari_comparison} shows that it only does so by a small amount, as the average difference in ARI is small.

\subsubsection*{Conclusion on clustering quality} 

From Figures \ref{fig:rank_comparison} and \ref{fig:ari_comparison} we conclude that COBRAS is the best choice in terms of clustering quality for the iterative clustering process that was outlined at the beginning of the paper.

\subsection*{Runtime}
Figure \ref{fig:runtimes} shows the ratio of the run time of COBRAS to the run times of its competitors for the 20 clustering tasks after performing 100 queries. It illustrates that COBRA is typically the fastest algorithm, which is not surprising as it requires only a single run of K-means. COBRAS requires multiple K-means runs, rendering it slower than COBRA. Compared to the other competitors, however, COBRAS is still fast. In particular, for the largest dataset COBRA still requires less than 10 seconds for the 100 queries, meaning that runtime will not be a limitation for the user while answering queries. MPCKMeans-NPU is significantly slower since it relies on a more expensive constraint-based variant of K-means, and requires re-clustering the entire dataset after every few queries. In contrast, COBRAS only re-clusters the parts of the dataset that are being refined. The high runtimes of COBS are explained by the fact that it generates a large number of unsupervised clusterings prior to querying the user. Once this set of clusterings is generated, however, selecting the clusterings is fast which means that COBS should not be disregarded for interactive settings.

\begin{sebcomment}
I would put Figures 3 and 4 into Figure 3 a) and b). This would clearly indicate that two have to be considered together.
\end{sebcomment}

\section{Conclusion}
We introduce COBRAS, an active clustering system based on the concept of super-instances. With its top-down strategy of constructing and refining super-instances it aims to produce high-quality clusterings in all stages of the clustering process. COBRAS is fast, since its most expensive step consists of performing K-means clustering on ever smaller parts of the data set. Our experiments confirm that COBRAS compares favorably to competitors in terms of both clustering quality and runtime, making it the preferred solution for constraint-based clustering in many settings.

\section*{Acknowledgements}
Toon Van Craenendonck is supported by the Agency for Innovation by Science and Technology in Flanders (IWT). This research is supported by Research Fund KU Leuven (GOA/13/010) and FWO (G079416N). This work has received funding from the European Research Council (ERC) under the European Union’s Horizon 2020 research and innovation programme (grant agreement No [694980] SYNTH: Synthesising Inductive Data Models).

\bibliographystyle{named}
\bibliography{references}

\begin{thebibliography}{}

\bibitem[\protect\citeauthoryear{Basu \bgroup \em et al.\egroup
  }{2004a}]{basu:sdm04}
Sugato Basu, Arindam Banerjee, and Raymond~J. Mooney.
\newblock Active semi-supervision for pairwise constrained clustering.
\newblock In {\em Proceedings of the 2004 SIAM International Conference on Data
  Mining (SDM-04)}, April 2004.

\bibitem[\protect\citeauthoryear{Basu \bgroup \em et al.\egroup
  }{2004b}]{probframework}
Sugato Basu, Misha Bilenko, and Raymond~J. Mooney.
\newblock A probabilistic framework for semi-supervised clustering.
\newblock In {\em Proceedings of the 10th ACM SIGKDD International Conference
  on Knowledge Discovery and Data Mining (KDD-2004)}, page 59–68, January
  2004.

\bibitem[\protect\citeauthoryear{Bilenko \bgroup \em et al.\egroup
  }{2004}]{Bilenko2004}
Mikhail Bilenko, Sugato Basu, and Raymond~J. Mooney.
\newblock Integrating constraints and metric learning in semi-supervised
  clustering.
\newblock In {\em Proc. of 21st International Conference on Machine Learning},
  pages 81--88, July 2004.

\bibitem[\protect\citeauthoryear{Caruana \bgroup \em et al.\egroup
  }{2006}]{Caruana06metaclustering}
Rich Caruana, Mohamed Elhawary, and Nam Nguyen.
\newblock {Meta clustering}.
\newblock In {\em Proc. of the International Conference on Data Mining}, 2006.

\bibitem[\protect\citeauthoryear{Davis \bgroup \em et al.\egroup
  }{2007}]{Davis:2007:IML:1273496.1273523}
Jason~V. Davis, Brian Kulis, Prateek Jain, Suvrit Sra, and Inderjit~S. Dhillon.
\newblock Information-theoretic metric learning.
\newblock In {\em Proceedings of the 24th International Conference on Machine
  Learning}, ICML '07, pages 209--216, New York, NY, USA, 2007. ACM.

\bibitem[\protect\citeauthoryear{Garc\'{i}a \bgroup \em et al.\egroup
  }{2010}]{GARCIA20102044}
Salvador Garc\'{i}a, Alberto Fern\'{a}ndez, Juli\'{a}n Luengo, and Francisco
  Herrera.
\newblock Advanced nonparametric tests for multiple comparisons in the design
  of experiments in computational intelligence and data mining: Experimental
  analysis of power.
\newblock {\em Information Sciences}, 180(10):2044 -- 2064, 2010.
\newblock Special Issue on Intelligent Distributed Information Systems.

\bibitem[\protect\citeauthoryear{Hodges and Lehmann}{1962}]{alignedrank}
J.~L. Hodges and E.~L. Lehmann.
\newblock Rank methods for combination of independent experiments in analysis
  of variance.
\newblock {\em The Annals of Mathematical Statistics}, 33(2):482--497, 1962.

\bibitem[\protect\citeauthoryear{Hubert and Arabie}{1985}]{ARI}
Lawrence Hubert and Phipps Arabie.
\newblock Comparing partitions.
\newblock {\em Journal of Classification}, 2(1):193--218, 1985.

\bibitem[\protect\citeauthoryear{Mallapragada \bgroup \em et al.\egroup
  }{2008}]{Mallapragada2008}
Pavan~K. Mallapragada, Rong Jin, and Anil~K. Jain.
\newblock Active query selection for semi-supervised clustering.
\newblock In {\em Proc. of the 19th International Conference on Pattern
  Recognition}, 2008.

\bibitem[\protect\citeauthoryear{Rangapuram and Hein}{2012}]{Rangapuram2012}
Syama~S. Rangapuram and Matthias Hein.
\newblock {Constrained 1-spectral clustering}.
\newblock In {\em Proc. of the 15th International Conference on Artificial
  Intelligence and Statistics}, 2012.

\bibitem[\protect\citeauthoryear{Szegedy \bgroup \em et al.\egroup
  }{2015}]{inceptionnet}
Christian Szegedy, Vincent Vanhoucke, Sergey Ioffe, Jonathon Shlens, and
  Zbigniew Wojna.
\newblock Rethinking the inception architecture for computer vision.
\newblock {\em CoRR}, abs/1512.00567, 2015.

\bibitem[\protect\citeauthoryear{Van~Craenendonck and Blockeel}{2017}]{COBS}
Toon Van~Craenendonck and Hendrik Blockeel.
\newblock Constraint-based clustering selection.
\newblock In {\em Machine Learning}. 2017.

\bibitem[\protect\citeauthoryear{Van~Craenendonck \bgroup \em et al.\egroup
  }{2017}]{COBRA}
Toon Van~Craenendonck, Sebastijan Duman\v{c}i\'{c}, and Hendrik Blockeel.
\newblock {COBRA: A Fast and Simple Method for Active Clustering with Pairwise
  Constraints}.
\newblock In {\em Proc. of the International Joint Conference on Artificial
  Intelligence}, 2017.

\bibitem[\protect\citeauthoryear{von Luxburg \bgroup \em et al.\egroup
  }{2014}]{ScienceOrArt}
Ulrike von Luxburg, Robert~C. Williamson, and Isabelle Guyon.
\newblock {Clustering: Science or Art?}
\newblock In {\em Workshop on Unsupervised Learning and Transfer Learning, JMLR
  Workshop and Conference Proceedings 27}, 2014.

\bibitem[\protect\citeauthoryear{Wagstaff \bgroup \em et al.\egroup
  }{2001}]{Wagstaff01constrainedk-means}
Kiri Wagstaff, Claire Cardie, Seth Rogers, and Stefan Schroedl.
\newblock {Constrained K-means Clustering with Background Knowledge}.
\newblock In {\em Proc. of the Eighteenth International Conference on Machine
  Learning}, pages 577--584, 2001.

\bibitem[\protect\citeauthoryear{Wang \bgroup \em et al.\egroup
  }{2014}]{Wang2014}
Xiang Wang, Buyue Qian, and Ian Davidson.
\newblock {On constrained spectral clustering and its applications}.
\newblock {\em Data Mining and Knowledge Discovery}, 28(1):1--30, 2014.

\bibitem[\protect\citeauthoryear{Xing \bgroup \em et al.\egroup
  }{2003}]{xing2002distance}
Eric~P. Xing, Andrew~Y. Ng, Michael~I. Jordan, and Stuart Russell.
\newblock Distance metric learning, with application to clustering with
  side-information.
\newblock In {\em Advances in Neural Information Processing Systems 15}, pages
  505--512, 2003.

\bibitem[\protect\citeauthoryear{Xiong \bgroup \em et al.\egroup
  }{2014}]{Xiong2014}
Sicheng Xiong, Javad Azimi, and Xiaoli~Z. Fern.
\newblock {Active learning of constraints for semi-supervised clustering}.
\newblock {\em IEEE Transactions on Knowledge and Data Engineering},
  26(1):43--54, 2014.

\end{thebibliography}

\end{document}